\documentclass{article}
\usepackage{spconf,amsmath,graphicx}
\usepackage{hyperref}
\usepackage{graphicx}
\usepackage{subcaption}
\usepackage{amsmath}
\usepackage{algorithm}
\usepackage{algorithmic}
\usepackage{booktabs}
\usepackage{url}

\title{Language Model is a Branch Predictor for Simultaneous Machine Translation}

\name{
Aoxiong Yin \sthanks{Both authors contributed equally to this research. † Corresponding author (siliang@zju.edu.cn). This work is supported by NSFC (No. 62272411), the National Key Research and Development Project of China (2018AAA0101900), and FinVolution Group.},
Tianyun Zhong\textsuperscript{*},
Haoyuan Li,
Siliang Tang\textsuperscript{†},
Zhou Zhao
}
\address{Zhejiang University}

\begin{document}
\maketitle
\begin{abstract}
  The primary objective of simultaneous machine translation (SiMT) is to minimize latency while preserving the quality of the final translation. Drawing inspiration from CPU branch prediction techniques, we propose incorporating branch prediction techniques in SiMT tasks to reduce translation latency. Specifically, we utilize a language model as a branch predictor to predict potential branch directions, namely, future source words. Subsequently, we utilize the predicted source words to decode the output in advance. When the actual source word deviates from the predicted source word, we use the real source word to decode the output again, replacing the predicted output. To further reduce computational costs, we share the parameters of the encoder and the branch predictor, and utilize a pre-trained language model for initialization. Our proposed method can be seamlessly integrated with any SiMT model. Extensive experimental results demonstrate that our approach can improve translation quality and latency at the same time. Our code is available at \url{https://github.com/YinAoXiong/simt_branch_predictor}.
\end{abstract}
\begin{keywords}
  simultaneous machine translation, language model, branch prediction
\end{keywords}
\section{Introduction}
\label{sec:intro}

Simultaneous machine translation (SiMT) \cite{choCanNeuralMachine2016,guLearningTranslateRealtime2017,maSTACLSimultaneousTranslation2019,arivazhaganMonotonicInfiniteLookback2019,maMonotonicMultiheadAttention2020} generates the target sequence as it receives the source sequence, instead of starting to generate the target sequence after reading the source sequence like full-sentence machine translation.
Most SiMT methods adopt a prefix-to-prefix architecture \cite{maSTACLSimultaneousTranslation2019}, which forces the target word to be aligned with the source prefix rather than the whole source sentence.
With this architecture, the model must determine which part of the target word should be aligned with the source prefix. If the source prefix is too extensive, the delay in translation will be excessive. Conversely, if the source prefix is too brief, there may not be sufficient information to accurately translate the target word.

However, from another perspective, the workflow of SiMT is like executing a serial program with branches in a loop, where the number of loops is the length of the source sequence, the number of branches is the size of the vocabulary, and the SiMT model and the above policy are the code executed in the branches.
Inspired by the CPU branch prediction technology \cite{smithStudyBranchPrediction1998}, we try to introduce a branch predictor in SiMT, so that we can improve the execution flow of SiMT from a higher dimension and reduce the translation delay.
Therefore, we introduce the language model as a branch predictor to predict the next possible execution branch: the next source word. It is worth noting that in the actual execution process, this prediction process can be executed in parallel with the translation process in the branch.
We will use the predicted output if the branch prediction is correct. If the branch prediction is wrong, we will withdraw the predicted output and re-translate the output using the real source words.

Since branch prediction is performed in a higher-dimensional outer layer, our method is transparent to the improvement of SiMT models, and it can be combined with any SiMT model \cite{elbayadEfficientWaitkModels2020,zhangModelingDualRead2022,zhangInformationTransportbasedPolicySimultaneous2022,zhangLearningAdaptiveSegmentation2020}.
In fact, our proposed method is also applicable to other modal simultaneous translation tasks, such as speec\cite{cheng2023mixspeech,cheng2023opensr}, vision\cite{yin2021simulslt,yin2022mlslt,yin2023gloss,lin2021simullr}, etc., as long as the input is discretized and the corresponding language model is constructed.
In addition, our method does not affect the final translation results due to the existence of the retraction mechanism.
Experimental results on two translation benchmarks, IWSLT15 en-vi \cite{iwslt2015} and WMT15 en-de \cite{bojar-EtAl:2015:WMT}, show that our method reduces latency while maintaining translation quality for both fixed \cite{maSTACLSimultaneousTranslation2019} and adaptive \cite{maMonotonicMultiheadAttention2020} SiMT policy. 
We also explore introducing pre-trained language models as shared encoders and branch predictors in SiMT, and the experimental results show that this method can significantly improve translation quality and latency.

\section{EVALUATION}
We adopt the evaluation framework proposed in \cite{DBLP:conf/icassp/ArivazhaganCTMB20}, which includes latency, stability, and quality metrics.
Translation quality is measured by calculating the BLEU \cite{papineniBleuMethodAutomatic2002} score of the final translated output.
Most latency metrics are based on a delay vector $g$, where $g_j$ reports how many source tokens were read before writing the $j^\mathit{th}$ target token.
Let $o_{i,j}$ be the $j^{\mathit{th}}$ token of the $i^{\mathit{th}}$ output
in a SiMT; $1 \leq i \leq I$ and $1 \leq j \leq J$.
For each position $j$ in the final output, we define $g_j$ as: 
\begin{equation}
g_j =  \min_i \textrm{ s.t. } {o}_{i', j'} = {o}_{I, j'} \; \forall i' \geq i \; \mathrm{and} \;  \forall j' \leq j
\end{equation}
that is, the number of source tokens read before the prefix ending in $j$ took on its final value.
Based on the delay vector $g$, we use the Average Lagging (AL) metric \cite{maMonotonicMultiheadAttention2020} to quantify translation latency.

\begin{equation}
  \label{eq:al}
  \mathrm{AL} = \frac{1}{J} \sum_{j=1}^J g_j - \frac{j-1}{J/I}
\end{equation}

To evaluate stability, we adopt the Average Withdrawal Rate (AWR), which is defined as $AWR=W/J$, where $W$ represents the number of withdrawals. In our work, AWR is equivalent to the NE metric proposed in \cite{DBLP:conf/icassp/ArivazhaganCTMB20}.

\section{Methodology}

\subsection{Branch Prediction for SiMT}
The branch prediction algorithm for SiMT is shown in Algorithm \ref{alg:BP}.
For the sake of convenience, we add a special token $\phi $ to the output vocabulary, which represents that the model decides to perform the READ action instead of the WRITE action for the current input $x_i$.
This special token does not participate in the calculation of the final translation quality and latency and is not used as input for the adaptive SiMT policy.
So the workflow of the traditional SiMT method before the end of the source sentence (we do not consider the autoregressive translation phase after the end of the source sentence, because the branch prediction technique is invalid in this phase) is as follows:
\begin{equation}
  y_j = f_{\theta}(x_{<i+1},y_{<j})
\end{equation}
where $f$ represents the SiMT model, $\theta$ represents the model parameters.

Unlike the traditional SiMT method, our branch prediction method uses a language model to perform the branch prediction task in parallel with the SiMT model performing the translation task.
Then the model uses the predicted $\widehat {x}_{i+1} $ translation output $y_{j+1}$ before the next source word $x_{i+1}$ arrives.
The formal representation is shown in lines 8-9 of Algorithm \ref{alg:BP}.
If the real next source word $x_{i+1}$ is equal to the predicted source word $\widehat{x}_{i+1}$, we will skip the translation and only perform the branch prediction operation, and directly use the predicted translation output $y_{j+1}$.
Because the input of the SiMT model is the same, it must have the same output, so we can directly use the predicted translation output $y_{j+1}$.
If they are not equal, we introduce a new action WITHDRAW, which represents replacing the previous output with the current output.
We will use the real next source word $x_{i+1}$ to translate, and then replace the previous predicted output.
The formal representation is shown in lines 4-5 of Algorithm \ref{alg:BP}.

\begin{algorithm}[t]
  \caption{Branch Prediction for SiMT. Because the SiMT model and the language model are independent, we compute line 4 to 8 in parallel.}
  \label{alg:BP}
  \begin{algorithmic}[1]
    \REQUIRE{$\mathbf{X}$ = source tokens, \\ $\;\;\;\;\;\;\;$ $ x_0 = BeginOfSequence $, \\ $\;\;\;\;\;\;\;$ $ y_0 = BeginOfSequence $,   \\  $\;\;\;\;\;\;\;$   $ i=1,j=1$.}
    \STATE $\widehat{x}_i\gets lm(x_{<i})$
    \STATE $y_j = f_{\theta}([x_{<i};\widehat{x}_i],y_{<j}) $
    \WHILE {$y_{j-1} \neq EndOfSequence$}
      \IF {$x_{i}\neq \widehat{x}_{i}$ \OR $x_i = EndOfSequence$  }
        \STATE $y_j = f_{\theta}(x_{<i+1};,y_{<j})$
      \ENDIF
      \IF {$x_{i}\neq EndOfSequence$}
        \STATE $\widehat{x}_{i+1} \gets lm(x_{x_{<i+1}})$
        \STATE $y_{j+1} \gets f_{\theta}([x_{<{i+1}};\widehat{x}_{i+1}],y_{<j+1})$
        \STATE $j\gets j+1$
      \ENDIF
      \STATE $i\gets i+1$
    \ENDWHILE
    \RETURN {target tokens $\mathbf{Y}$}
  \end{algorithmic}
\end{algorithm}

\subsection{Shared Encoder and Language Model}
\label{sec:share}
To further save resources, we try to introduce a pre-trained language model (here GPT2) \cite{Radford2019LanguageMA} as a shared encoder and branch predictor in SiMT, and add an auxiliary language model loss (LM Loss) after the encoder.
The language model can introduce prior knowledge similar to that used by a human simultaneous translator when translating \cite{vandepitte2001anticipation}. For example, when he sees that the speaker has said "I love", he will guess that he may say "you" with a high probability.
In addition, inspired by \cite{chenMakingMostCrossLingual2022}, we first froze the parameters of the encoder. Then we unfroze the encoder after the randomly initialized decoder was sufficiently trained to train the whole model.
So the loss function of the first stage is as follows:

\begin{equation}
    \mathcal{L}_{\theta_{dec}} =\sum_{\langle  \mathbf{X},\mathbf{Y} \rangle \in D } \log(p(\mathbf{Y}|\mathbf{X};,\theta_{dec}))
\end{equation}
where $D$ represents the training set, and $\theta_{dec}$ represents the parameters of the decoder.

And when the decoder is sufficiently trained, the loss function of the second stage is as follows:

\begin{equation}
  \begin{split}
    \mathcal{L}_{\theta} =&\sum_{\langle  \mathbf{X},\mathbf{Y} \rangle \in D } \log(p(\mathbf{Y}|\mathbf{X};,\theta)) + \\ &\lambda \sum_{  \mathbf{X}\in D } \sum_{i=1}^{N} \log(p(x_i|x_{<i};,\theta_{enc}))
  \end{split}
  \label{eq:loss}
\end{equation}

where $\theta$ represents the parameters of the whole model, $\theta_{enc}$ represents the parameters of the encoder, and $\lambda$ represents the weight of the language model loss.

\begin{figure*}[t]
  \setlength{\abovecaptionskip}{2pt}
  \subcaptionbox{en $\rightarrow $ vi}{\includegraphics[width = 0.28\linewidth]{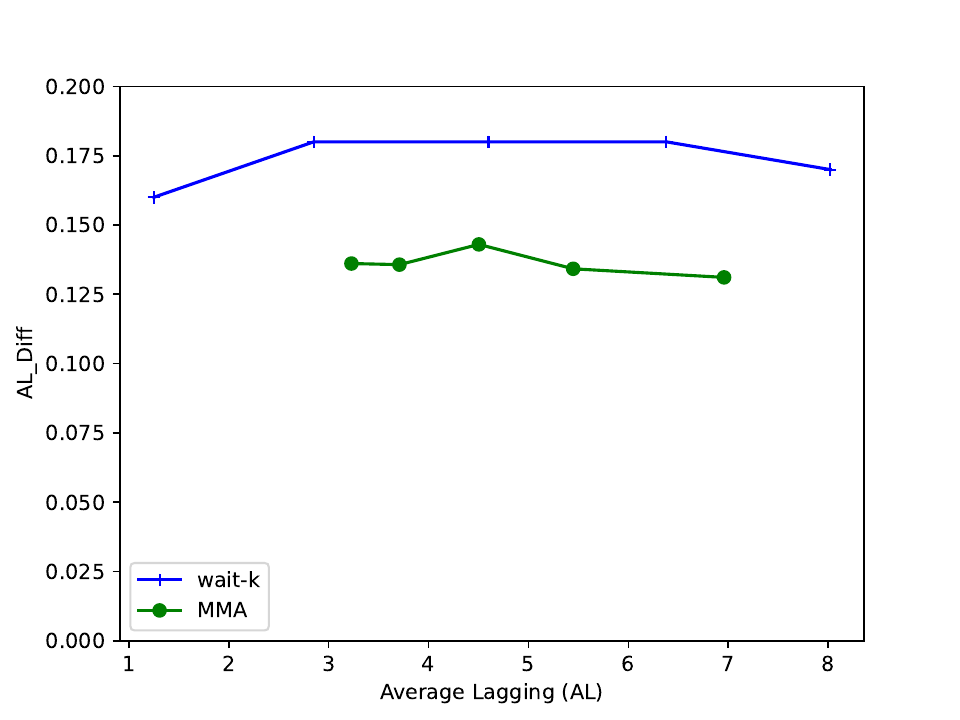}}
  \subcaptionbox{en $\rightarrow $ de }{\includegraphics[width = 0.28\linewidth]{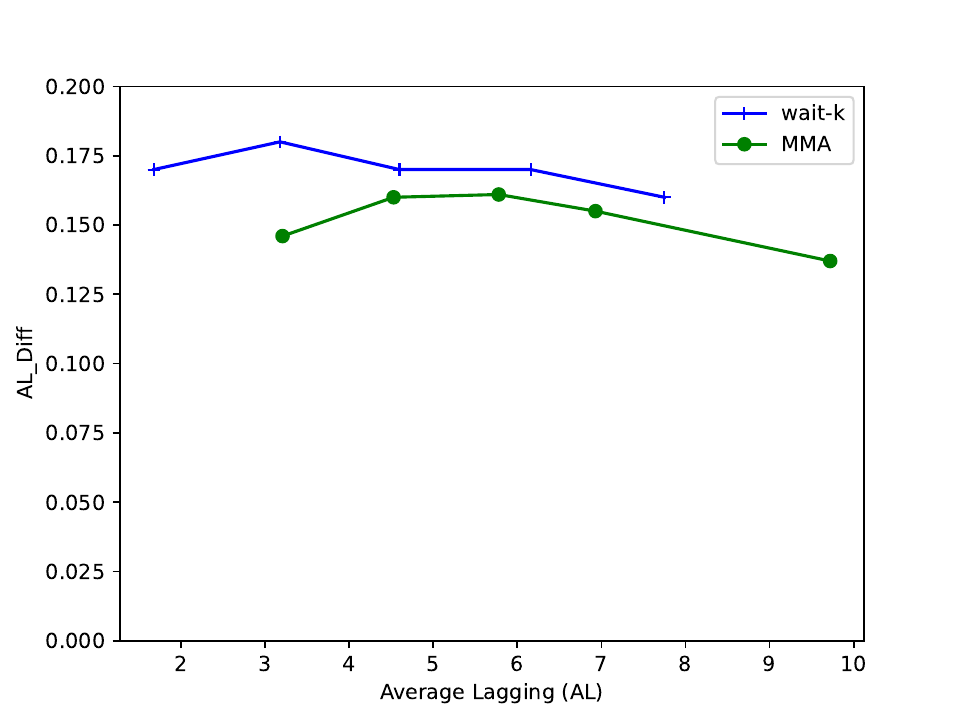}}
  \subcaptionbox{de $\rightarrow $ en}{\includegraphics[width = 0.28\linewidth]{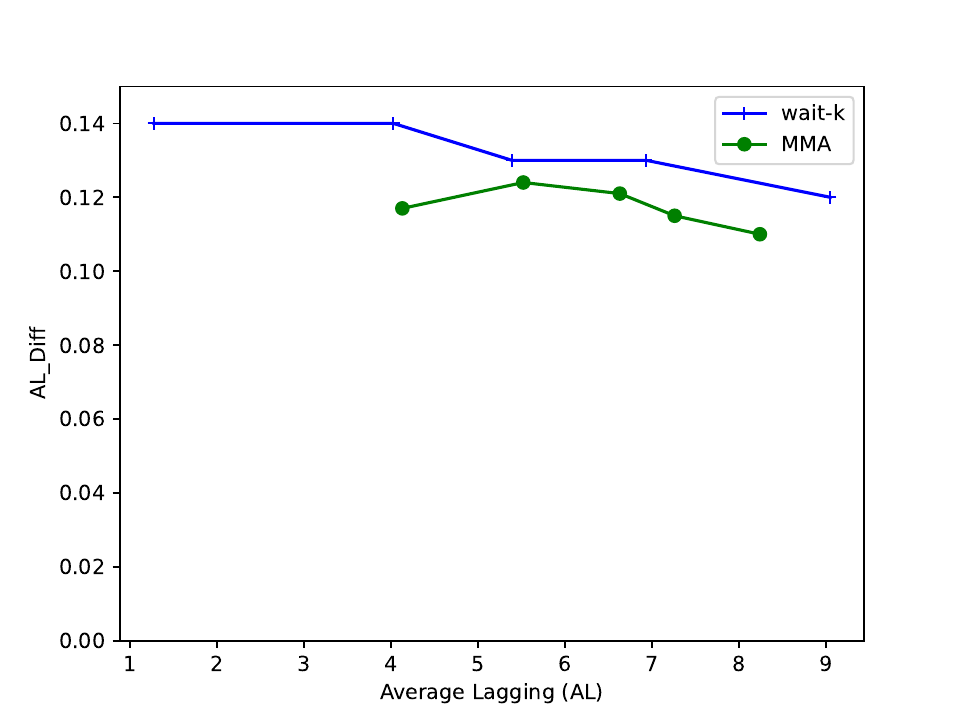}}
  \centering
  \caption{ Latency performance improvement for different language directions and different original AL cases after using branch prediction technology, in the case of $AWR < 0.7$.   The horizontal axis represents the original model's AL, and the vertical axis represents the difference between the AL after using branch prediction technology and the original model's AL.}
  \label{fig:main_result}
  \vspace{-5mm}
\end{figure*}

\section{Experiments}
\subsection{Datasets and Settings}
We evaluated the effectiveness of our proposed branch prediction method on two datasets: 1) \textbf{IWSLT15 English$\rightarrow $Vietnamese (En$\rightarrow $Vi)} 2) \textbf{WMT15 German$\leftrightarrow $English (De$\leftrightarrow $En)}. 
To ensure consistency with prior work, we adopted the same data processing approach as described in \cite{maMonotonicMultiheadAttention2020}.
We conduct experiments on following systems. \textbf{1) Wait-k} 
 \cite{maSTACLSimultaneousTranslation2019} is one of the most representative fixed SiMT policy, which first waits for k source words, then translates a target word and waits for a source word alternately.
\textbf{2) Monotonic multi-head attention (MMA)}
 \cite{maMonotonicMultiheadAttention2020} is one of the most commonly used adaptive SiMT policy. Every time it reads a source word, it predicts a Bernoulli random variable according to the source prefix and target prefix to decide whether to READ or WRITE.
\textbf{3) +BP} Apply the proposed branch prediction technique on wait-k or MMA.

\subsection{Branch Prediction}
We use the AL difference (the y-axis in Figure \ref{fig:main_result}) to evaluate the latency performance improvement after adding the branch prediction technique.
The formal expression is as follows:

\begin{equation}
  AL\_diff = AL(\text{baseline}) - AL(\text{+BP})
\end{equation}

Figure \ref{fig:main_result} shows that our designed branch prediction technique can bring latency performance improvement for different language directions SiMT.
Furthermore, it can be noticed that our proposed method is effective for both fixed and adaptive policy of wait-k \cite{maSTACLSimultaneousTranslation2019} and MMA \cite{maMonotonicMultiheadAttention2020}.

\begin{figure}[htbp]
  \centering
  \includegraphics[width=0.6\linewidth]{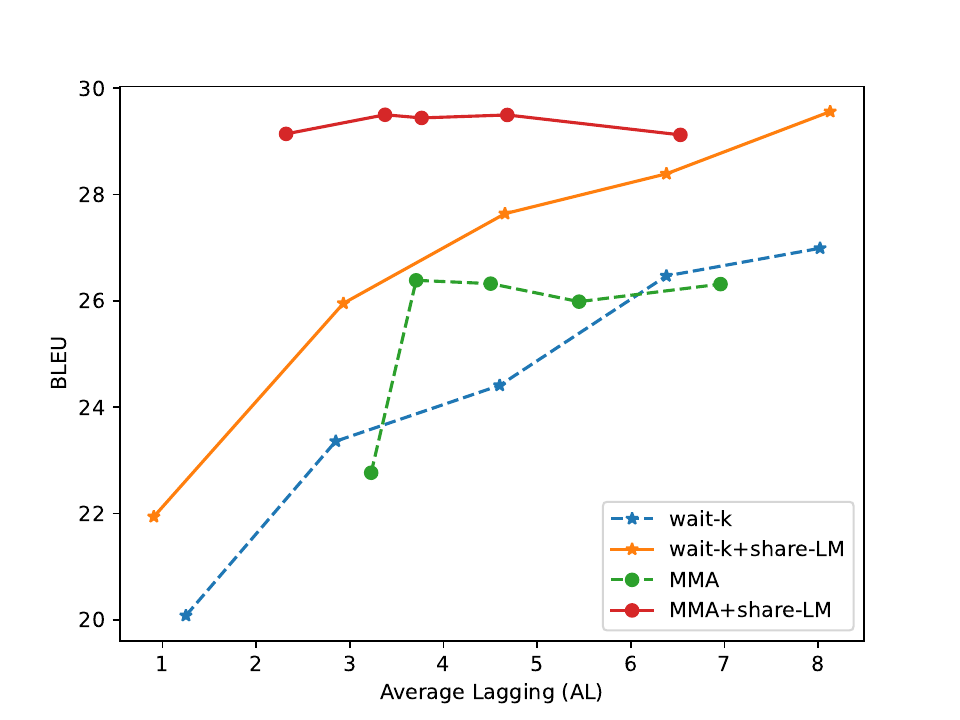}
  \caption{  The performance improvement brought by introducing the pre-trained language model and LM loss proposed in Section \ref{sec:share} in the en$\rightarrow$vi direction.}
  \label{fig:share_result}
  \vspace{-5mm}
\end{figure}

\subsection{Shared Encoder and Language Model}
First, we verify whether introducing the pre-trained language model and LM loss proposed in Section \ref{sec:share} can improve the model's performance on translation quality and latency.
The experimental results are shown in Figure \ref{fig:share_result}, which shows that introducing the pre-trained language model and LM loss significantly improves both fixed and adaptive policy of SiMT.
It can be seen that our method is better than the existing method in both translation quality and latency.

\begin{table}[htbp]

  \caption{The delay improvement performance of using the shared encoder as the branch predictor on the en$\rightarrow$vi direction wait-k \cite{maSTACLSimultaneousTranslation2019} method.}
  \label{tab:share_encoder_wiat_k}
  \center
  \resizebox{0.8\linewidth}{!}{
  \begin{tabular}{lcccccc}
    \toprule
    Model& \textbf{K=1} &\textbf{ K=3} & \textbf{K=5} & \textbf{K=7} & \textbf{K=9} \\
    \midrule
      GPT2-small & 0.157 & 0.179 &0.131 & 0.176 & 0.166 \\
      \midrule
      GPT2-small-\\share-encoder & \textbf{0.210}  & \textbf{0.225} & \textbf{0.248} & \textbf{0.211} & \textbf{0.198} \\
    \bottomrule
    \end{tabular}
  }
\end{table}

\begin{table}[htbp]

  \caption{The delay improvement performance of using the shared encoder as the branch predictor on the en$\rightarrow$vi direction MMA \cite{maMonotonicMultiheadAttention2020} method. Where L represents the weight of the delay loss.}
  \label{tab:share_encoder_MMA}
  \center
  \resizebox{0.8\linewidth}{!}{
  \begin{tabular}{lcccccc}
    \toprule
    Model& \textbf{L=0.5} &\textbf{ L=0.2} & \textbf{L=0.1} & \textbf{L=0.05} & \textbf{L=0.02} \\
    \midrule
      GPT2-small & 0.131 & 0.135 &0.131 & 0.135 & 0.132 \\
      \midrule
      GPT2-small-\\share-encoder & \textbf{0.248}  & \textbf{0.265} & \textbf{0.264} & \textbf{0.255} & \textbf{0.251} \\
    \bottomrule
    \end{tabular}
  }
\end{table}

Then, we experimentally verify whether using a shared encoder as a branch predictor can achieve the same effect as using an independent language model with the same parameters.
The experimental results are shown in Table \ref{tab:share_encoder_wiat_k} and Table \ref{tab:share_encoder_MMA}, which shows that compared with using an independent language model, using a shared encoder as a branch predictor can even achieve better results.
One possible reason why the branch prediction effect is better is that the addition of the LM loss is equivalent to fine-tuning the language model using the training data.
The subsequent fine-tuning analysis experiment results support this idea.

\subsection{Analysis}
Next, we continue to conduct a wide range of analyses to understand the specific factors that affect branch prediction.
Unless otherwise stated, all results are reported on en$\rightarrow$vi.

\begin{figure}[htbp]
  \subcaptionbox{en $\rightarrow $ vi wait-k}{\includegraphics[width = 0.45\linewidth]{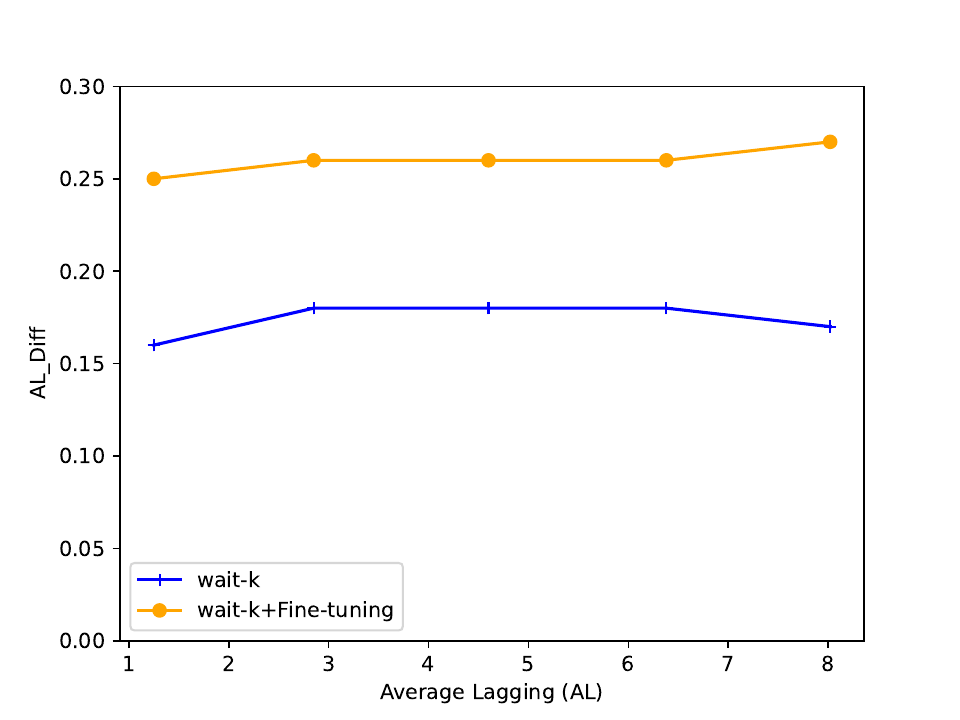}}
  \hfill
  \subcaptionbox{en $\rightarrow $ de MMA}{\includegraphics[width = 0.45\linewidth]{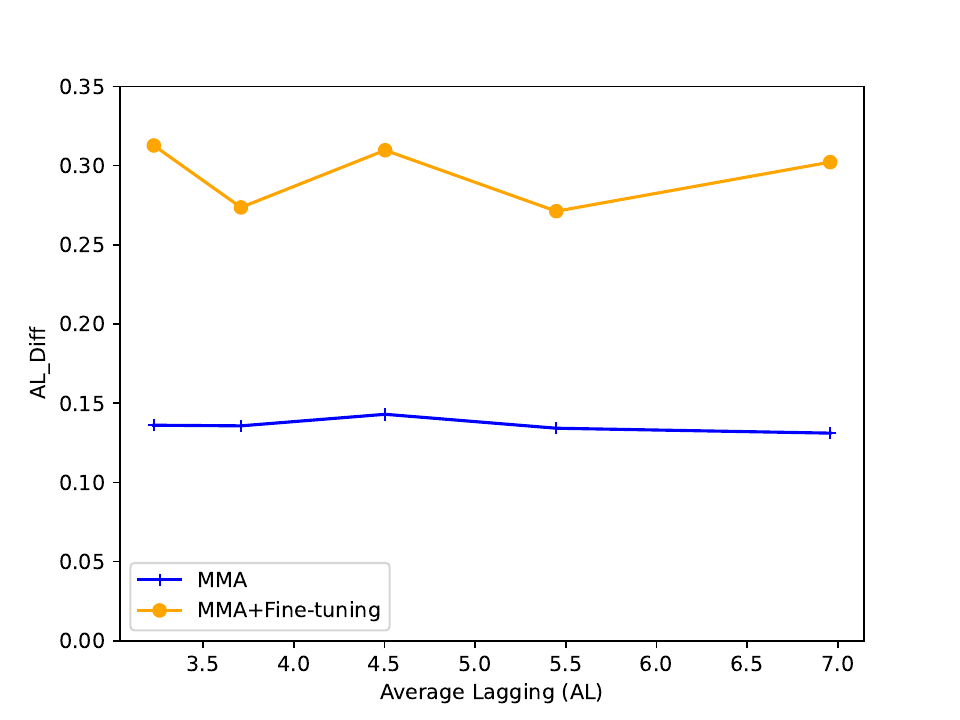}}
  \centering
  \vspace{-3mm}
  \caption{ Comparison of fine-tuning the language model and using the original language model for SiMT latency performance improvement.}
  \label{fig:ft}
  \vspace{-3mm}
\end{figure}

\begin{table}[htbp]

  \caption{Analyze the impact of using different language model sizes as branch predictors on latency improvement. Experiments are conducted on the en$\rightarrow$de direction wait-K method.}
  \label{tab:gpt_size}
  \center
  \resizebox{0.8\linewidth}{!}{
  \begin{tabular}{lcccccc}
    \toprule
    Model& \textbf{K=1} &\textbf{ K=3} & \textbf{K=5} & \textbf{K=7} & \textbf{K=9} \\
    \midrule
      GPT2-small & 0.162     & 0.171 &0.169 & 0.162 & 0.154 \\
      \midrule
      GPT2-medium & 0.174    & 0.184 &0.185 & 0.175& 0.166 \\
      \midrule
      GPT2-large & 0.174     & 0.190 &0.187 & 0.178 & 0.168 \\
      \midrule
      GPT2-xl    &0.184      & 0.195 &0.192 & 0.183 & 0.179 \\
      \midrule
      GPT2-small-\\fine-tuning & \textbf{0.212}  & \textbf{0.224} & \textbf{0.227} & \textbf{0.223} & \textbf{0.218} \\
    \bottomrule
    \end{tabular}
  }
\end{table}

\textbf{Fine-tuning the Language Model}
The accuracy of the branch predictor is pivotal to the success of branch prediction techniques. However, pre-trained language models trained on internet-based monolingual data may exhibit suboptimal performance when applied to specific tasks. Fine-tuning the language model using task-specific training data presents a natural solution for improving its predictive capacity. The efficacy of fine-tuning is demonstrated in Figure \ref{fig:ft}, where the fine-tuned branch predictor outperforms the original model by 49.6\% and 116.1\% on wait-K and MMA tasks, respectively. These results highlight the effectiveness of fine-tuning in enhancing the predictive accuracy of the branch predictor.

\textbf{Language Models of Different Sizes}
We also investigated the performance of the model using pre-trained GPTs of different sizes as branch predictors, and the results are shown in Table \ref{tab:gpt_size}. It can be seen that using a larger pre-trained GPT can bring improvements, but the cost performance is not as good as fine-tuning a small GPT model.

\begin{table}[htb]
  \centering
  \small
  \caption{Ablation experiments on the model with a delay loss weight of 0.1 for the MMA method.}
  \label{tab:ab_share}
  \begin{tabular}{lcc}
    \toprule  %
    Model&BLEU$\uparrow $&AL$\downarrow $\\
    \midrule  %
    MMA     &26.97 	&4.55\\
    \midrule
    +Pre-trained LM     &29.36 	&4.19  \\
    \midrule
    +LM loss         &27.07 	&4.86 \\
    \midrule
    +Pre-trained LM\\+LM loss &\textbf{29.44} 	&\textbf{3.76}\\
    \bottomrule %
    \end{tabular}
\end{table}

\textbf{Pre-trained Language Model and LM Loss}
We conducted a ablation study on the proposed pre-trained language model and LM loss in Section \ref{sec:share}.
The experimental results are shown in Table \ref{tab:ab_share}, which shows that the pre-trained language model and LM loss can both improve the translation quality and delay of the model, and the combination of the two achieves the best results.

\begin{figure}[htbp]
  \centering
  \includegraphics[width=0.6\linewidth]{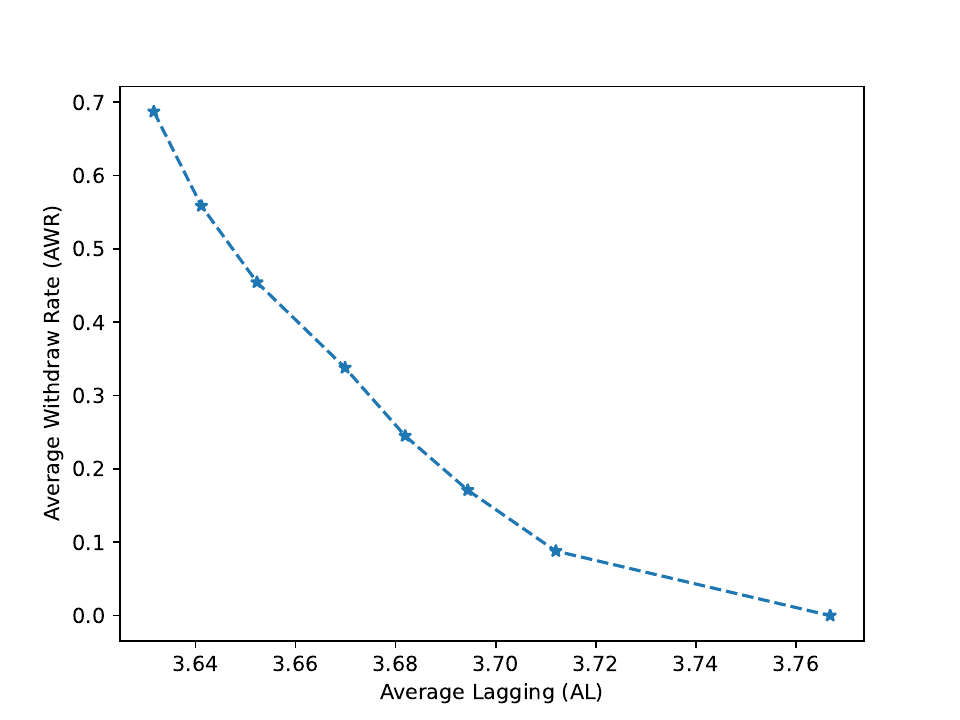}
  \caption{Model performance using different branch prediction probability thresholds for the MMA model with latency loss weight = 0.1.}
  \label{fig:awr}
  \vspace{-5mm}
\end{figure}

\textbf{Different Probability Thresholds}
The branch predictor can trade off between $AWR$ and $AL_diff$ by adjusting the decision probability threshold. A test result is shown in Figure \ref{fig:awr}.
\section{Conclusion}
In this paper, we propose branch prediction techniques for SiMT and introduce pre-trained language models for the first time into SiMT.
Experiments show that our branch prediction method is effective for both fixed and adaptive policy models and can effectively reduce translation latency.
In addition, the introduction of pre-trained language models can significantly improve the translation quality and latency of the model.

\newpage
\bibliographystyle{IEEEbib}
\bibliography{strings,refs,acl2023_references,custom,cr_reference.bib}

\end{document}